\documentclass[10pt,twocolumn,letterpaper]{article}
\usepackage[utf8]{inputenc}

\usepackage[pagenumbers]{cvpr}      % To produce the REVIEW version
\definecolor{cvprblue}{rgb}{0.21,0.49,0.74}
\usepackage[pagebackref,breaklinks,colorlinks,allcolors=cvprblue]{hyperref}

\title{MixAR: Mixture Autoregressive Image Generation}

\author{%
    Jinyuan Hu\textsuperscript{1,2*}
    \qquad
    Jiayou Zhang\textsuperscript{2}
    \qquad
    Shaobo Cui\textsuperscript{3}
    \qquad
    Kun Zhang\textsuperscript{2,4}
    \qquad
    Guangyi Chen\textsuperscript{2,4}
    \vspace{4mm}
    \\
    \begin{minipage}{\textwidth}
        \centering
        \small
        \textsuperscript{1}Tsinghua University \\[0.5ex]
        \textsuperscript{2}Mohamed bin Zayed University of Artificial Intelligence (MBZUAI) \\[0.5ex]
        \textsuperscript{3}\'Ecole Polytechnique F\'ed\'erale de Lausanne (EPFL) \\[0.5ex]
        \textsuperscript{4}Carnegie Mellon University
    \end{minipage}
}

\begin{document}
\maketitle

\renewcommand{\thefootnote}{*}
\footnotetext{This work was done during Jinyuan Hu's visit to Mohamed bin Zayed University of Artificial Intelligence (MBZUAI).}
\renewcommand{\thefootnote}{\arabic{footnote}}

\begin{abstract}

% Discrete autoregressive (AR) approaches, which represent images as sequences of tokens from a finite codebook, have achieved remarkable success in image generation. 

Autoregressive (AR) approaches, which represent images as sequences of discrete tokens from a finite codebook, have achieved remarkable success in image generation. 
However, the quantization process and the limited codebook size inevitably discard fine-grained information, placing bottleneck on fidelity. Motivated by this limitation, recent studies have explored autoregressive modeling in continuous latent spaces, which offers higher generation quality. 
Yet, unlike discrete tokens constrained by a fixed codebook, continuous representations lie in a vast and unstructured space, posing significant challenges for efficient autoregressive modeling. 
To address these challenges, we introduce \textbf{MixAR}, a novel framework that leverages mixture training paradigms to inject discrete tokens as prior guidance for continuous AR modeling. 
MixAR is a factorized formulation that leverages discrete tokens as prior guidance for continuous autoregressive prediction. We investigate several discrete–continuous mixture strategies, including self-attention (DC-SA), cross-attention (DC-CA), and a simple approach (DC-Mix) that replaces homogeneous mask tokens with informative discrete counterparts.
Moreover, to bridge the gap between ground-truth training tokens and inference tokens produced by the pre-trained AR model, we propose Training–Inference Mixture ({TI-Mix}) to achieve consistent training and generation distributions.
In our experiments, we demonstrate a favorable balance of the DC-Mix strategy between computational efficiency and generation fidelity, and consistent improvement of TI-Mix.

% making the "specificially" more detialed.
% In our experiments, we demonstrate that our MixAR achieves a favorable balance between computational efficiency and generation fidelity.
% In our experiments, we demonstrate that the simple \textbf{MixAR} paradigm outperforms other hybrid strategies, such as using discrete tokens as a prefix sequence for self- or cross-attention.
%highlight the concrete part of the experiments

\end{abstract}     
\section{Introduction}
\label{sec:intro}

% ---------- Figure 1: DC-Mix Visual ----------
\begin{figure}[t]
    \centering
    \begin{subfigure}[t]{0.48\linewidth}
        \includegraphics[width=\linewidth]{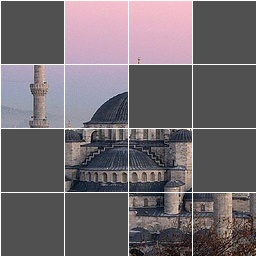}
        \caption{MAR}
    \end{subfigure}
    \hfill
    \begin{subfigure}[t]{0.48\linewidth}
        \includegraphics[width=\linewidth]{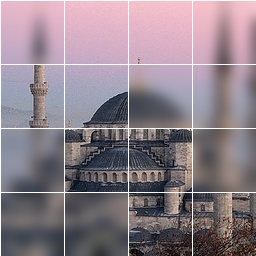}
        \caption{MixAR}
    \end{subfigure}
    \vspace{1mm}
    \caption{
        % \textbf{Intuition behind MixAR}. Compared with the traditional masked autoregressive (\textbf{MAR}) strategy—which can be intuitively understood as predicting the full image from only a subset of visible patches, as illustrated in (a)—\textbf{MixAR} reconstructs the image from a partially degraded yet informative version, as shown in (b). This shift provides richer contextual cues and reflects the core idea behind our approach.
        \textbf{Intuition behind MixAR}. Compared with the traditional masked autoregressive (MAR) strategy that predicts the full image from only a subset of visible patches (a), MixAR reconstructs the image from a partially degraded yet still informative version (b). This design provides richer contextual cues and captures fine-grained details more effectively.
    }
    \label{fig:dcmix_visual}
    \vspace{-3mm}
\end{figure}

Recent successes in large language models (LLMs) have demonstrated the remarkable capability of the  autoregressive framework in sequence modeling~\cite{deepseek,gpt4,llama3,qwen}. Motivated by this success, a growing line of research has adapted AR modeling to image generation~\cite{igpt,vqgan,llamagen,maskgit,mage,rar,maskbit}. Mainstream AR methods use discrete-valued tokens as input contexts and generation targets, as discrete modeling can be efficiently optimized, such as using the cross-entropy loss. However, to compress an image into a sequence of discrete tokens, the quantization step is indispensable, which maps continuous latent vectors to discrete codewords in a finite codebook~\cite{vqvae,rqvae,lfq,fsq,bsq}, inevitably causing information loss. Moreover, the representational capacity of these models is fundamentally limited by the codebook size, while enlarging the codebook often leads to optimization difficulties. 
These constraints impose an intrinsic upper bound on the fidelity achievable by discrete autoregressive models. 

In contrast, recent studies have explored continuous autoregressive models that directly learn the joint distribution of continuous latents without quantization~\cite{mar}. Despite their conceptual appeal, these methods face optimization challenges, as continuous representations reside in a vast and unstructured space, unlike their discrete counterparts constrained by a codebook. This creates a fundamental dilemma: discrete models are easier to train but inherently lossy, while continuous models offer higher fidelity yet suffer from training inefficiency.

\begin{figure*}[t]
    \centering
    \includegraphics[width=\linewidth,trim=5 0 10 4,clip]{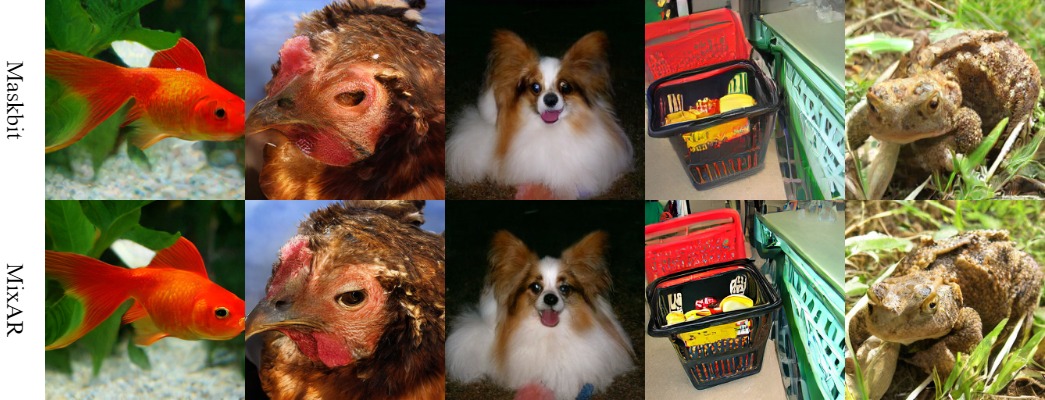}
    \caption{
        \textbf{MixAR vs Maskbit}
        Top: Images generated by Maskbit~\cite{maskbit}.
        Bottom: Images generated by our MixAR-L model conditioned on the discrete tokens produced by Maskbit, demonstrating significant improvements in generation quality.
    }
    \vspace{-2mm}
    \label{fig:compare}
\end{figure*}

To address this challenge, we take inspiration from classic continuous regression methods~\cite{sun2013deep,zhang2014facial}, which mitigate optimization difficulties by performing a coarse discrete classification prior to continuous regression. This cascaded refinement paradigm has also been proven effective in other vision tasks like object detection~\cite{ren2015faster} and pose estimation~\cite{cao2017realtime}. We hypothesize that a similar factorization can also benefit image generation. Specifically, rather than directly modeling the continuous latent distribution in a single stage, we propose to decouple the process into two complementary components: a discrete module that first predicts discrete tokens (formulated as a classification problem), and a continuous module that generates high-fidelity continuous latents under the guidance of these classification results (formulated as a regression problem). This factorization combines the stability of discrete training with the expressive power of continuous modeling, resolving the trade-off between optimization efficiency and fidelity.

Motivated by this insight, we propose \textbf{MixAR}, a novel framework that leverages a mixture of discrete and continuous tokens to incorporate guidance from discrete representations. Instead of directly learning the joint distribution of continuous latents~\cite{mar}, MixAR adopts factorized formulation that models the conditional continuous distribution given discrete variables. 
In this formulation, off-the-shelf discrete AR models are employed to capture the distribution over discrete tokens, while the continuous component is learned through a masked autoregressive strategy guided by the discrete variables. As shown in Figure~\ref{fig:dcmix_visual}, the standard MAR framework replaces masked regions with a single, non-informative mask token, predicting the full image from visible patches only. In contrast, DC-Mix substitutes these regions with discrete tokens that preserve partial semantic and structural information, enabling reconstruction from a partially degraded input rather than entirely missing regions.

Furthermore, we investigate several strategies for injecting discrete guidance into the continuous AR model, including self-attention (DC-SA), cross-attention (DC-CA), and a direct mixture approach (DC-Mix). Specifically, DC-SA and DC-CA utilize discrete tokens as prefix inputs and incorporate their guidance into continuous modeling through self- and cross-attention mechanisms, respectively. DC-Mix, rather than adding prefix discrete tokens, directly replaces homogeneous, non-informative mask tokens with semantically rich discrete context. Interestingly, we found that DC-Mix introduces no additional parameters yet achieves fidelity comparable to methods using prefix tokens.
In practice, if absolute performance is the priority and a higher training cost is acceptable, DC-SA is the most effective choice, as it utilizes all tokens for guidance. For computationally constrained settings, we recommend DC-Mix, which provides a favorable trade-off between efficiency and fidelity.

In addition, we find that incorporating discrete guidance inevitably introduces a distribution mismatch between the ground-truth discrete tokens used during training and the model-generated tokens encountered at inference.
% This gap is inherent to all guidance strategies—whether the discrete tokens are introduced through self-attention prefixing, cross-attention conditioning, or our proposed DC-Mix mechanism. 
To address this universal issue, we introduce Training–Inference Mixture ({TI-Mix}), a simple yet broadly applicable plug-in technique that gradually interpolates between ground-truth and generated discrete tokens during training. Specifically, we jointly mix the continuous contextual tokens, ground-truth discrete tokens, and generated discrete tokens from the pre-trained discrete AR model. During training, we progressively replace ground-truth priors with generated ones, allowing the continuous AR model to gradually adapt to realistic inference conditions.

Through extensive experiments, we demonstrate that MixAR outperforms the standard MAR consistently on both model sizes. By systematically analyzing different guidance strategies, we find that DC-Mix achieves an attractive balance between computational efficiency and generation fidelity, while TI-Mix further improves generation quality by reducing the training–inference discrepancy. As illustrated in Figure~\ref{fig:compare}, MixAR produces substantially more detailed and coherent images than MaskBit, highlighting the advantage of leveraging discrete priors to guide continuous latent prediction.

In summary, our contributions are as follows:
\begin{itemize}
    \item We reformulate continuous autoregressive modeling as a \textbf{factorized process}, which incorporates the priors of discrete tokens into the continuous space, achieving better optimization efficiency and generation fidelity. 
    \item We systematically compare different guidance strategies, such as DC-SA and DC-CA, and \textbf{DC-Mix}, and find that the direct mixture of DC-Mix achieves a favorable trade-off between efficiency and fidelity.
    \item We introduce \textbf{TI-Mix} to gradually blend ground-truth and generated discrete tokens during training, reducing the distribution mismatch between training and inference. 
\end{itemize}

\section{Related Work}
\label{sec:related work}

\subsection{Discrete Autoregressive Image Generation}

Discrete autoregressive modeling has been widely adopted for image generation~\cite{igpt,vqgan,llamagen,maskgit,haltonmaskgit,mage,maskbit}, inspired by the advances of large language models in sequence modeling.  
These methods quantize images into sequences of discrete tokens (e.g., via VQ-VAE~\cite{vqvae} or VQ-GAN~\cite{vqgan}) and model their joint distribution using cross-entropy loss.  
Existing approaches can be broadly divided into two paradigms: the next-token prediction paradigm and the masked autoregressive paradigm, depending on how conditional dependencies are formulated and tokens are generated during inference.

The next-token paradigm, derived from GPT-style language modeling~\cite{gpt3}, predicts tokens sequentially~\cite{igpt, vqgan, llamagen}, while the masked autoregressive paradigm, inspired by BERT-style masked prediction~\cite{bert}, predicts multiple masked tokens in parallel, generating a subset at each step until all tokens are produced~\cite{maskgit, haltonmaskgit, mage, maskbit}.  
Despite their differences, both paradigms fundamentally rely on quantization~\cite{vqvae,vqgan,lfq,fsq,bsq} to obtain discrete representations, making them subject to the information loss and limited representational capacity imposed by finite codebooks.  
These constraints motivate the exploration of continuous autoregressive modeling, which seeks to bypass quantization by directly learning in continuous latent spaces.

\subsection{Continuous Autoregressive Image Generation}

Continuous autoregressive (AR) modeling aims to bypass the quantization bottleneck of discrete methods by directly learning dependencies in continuous latent spaces.
The pioneering work MAR~\cite{mar} establishes the first strong continuous AR baseline by applying masked autoregressive modeling to continuous latents and refining predictions with a diffusion-based denoising head.
By eliminating codebook quantization, MAR demonstrates the potential of continuous representations to achieve higher fidelity and richer generative capacity.
Following MAR, several subsequent works explore alternative formulations of continuous autoregression. For example, FlowAR~\cite{flowar} incorporates flow-based parameterizations to better capture complex latent distributions, while HART~\cite{hart} leverages hybrid transformer–diffusion architectures to improve stability and scalability.
Although these methods introduce valuable design choices, continuous AR models as a whole remain challenging to optimize due to the vast, unstructured nature of continuous latent spaces. Our motivation is that the expressive potential of continuous representations can be more effectively realized through a factorized formulation that introduces discrete priors, along with mixture strategies that provide more stable and informative guidance during autoregressive prediction.

\renewcommand{\subsubsectionautorefname}{Section}
% ---------- Figure: Pipeline ----------
\begin{figure*}[t]
    \centering
    \includegraphics[width=\textwidth,trim=10 5 10 5,clip]{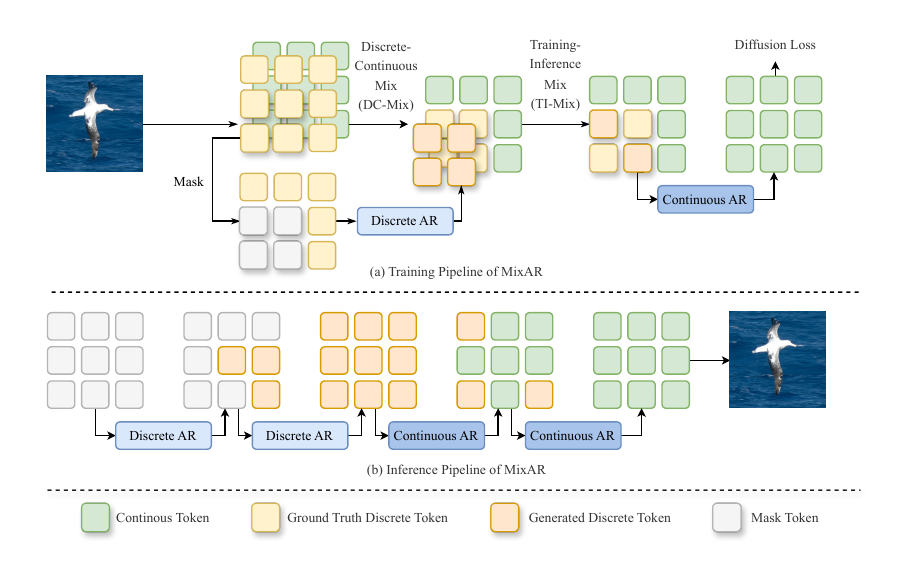}
    \caption{
        \textbf{Overview of the training and inference pipeline of MixAR.} During training, we first mix ground truth continuous and discrete tokens with DC-Mix, followed by TI-Mix that replaces a certain portion of ground truth discrete tokens with generated ones. The continuous AR model then predicts the ground-truth continuous tokens given the mixed tokens, optimized by diffusion loss.
        During inference, we first generate discrete tokens with a pretrained autoregressive model and use them to guide the continuous autoregressive model to generate more information-rich continuous tokens. 
    }
    \label{fig:pipeline}
\end{figure*}

% \jiayou{adding the full name (TI-Mix = Training-Inference Mixture) in the figure helps reading. I think the name "MixAR" should occur in the figure as well, because we need to emphasize that MixAR is a strategy containing both DC-Mix and TI-Mix and covers both training and test time.}

\section{Method}
\label{sec:method}
% \subsection{Preliminaries}

In this section, we first present the preliminaries of continuous autoregressive modeling (MAR). We then illustrate how to leverage factorization to decompose continuous modeling into a discrete autoregressive component and a conditional module that generates continuous tokens based on discrete ones. Finally, we discuss how to inject discrete tokens as guidance for continuous generation and how to bridge the gap between ground-truth training tokens and inference tokens produced by the pre-trained AR model.

\subsection{Masked Autoregressive Model (MAR)}
\label{sec:mar}

\paragraph{Tokenization.}
Given an image $I \in \mathbb{R}^{H\times W\times 3}$, the discrete AR models first utilize a pre-trained tokenizer such as VQVAE to encode $I$ into latent representations $I'\in\mathbb{R}^{h\times w\times d}$, where $h,w,d$ denote the latent height, width, and channel dimensions respectively.  
The latent feature map $I'$ is then reshaped into a sequence of $N=h\times w$ latents
\[
X_d = [x_d^1, x_d^2, \ldots, x_d^N],
% X_c = [x_c^1, x_c^2, \ldots, x_c^N],
\]
where each $x_d^i \in \mathbb{R}^d$ is a discrete token.  

\paragraph{Autoregessive modeling.}
The joint probability of all tokens is then factorized autoregressively as
\begin{equation}
    p(\mathbf{x}_d) = \prod_{i=1}^{N} p(x_d^i \mid x_d^{<i}),
\end{equation}
and are typically trained using the cross-entropy loss:
\begin{equation}
    \mathcal{L}_{\text{AR}}
    = -\sum_{i=1}^{N} \log p_\theta(x_d^i \mid x_d^{<i}).
\end{equation}
While next-token prediction models effectively capture long-range dependencies, the strict autoregressive ordering limits flexibility in modeling two-dimensional spatial relationships.

\paragraph{Masked autoregressive paradigm.}
In contrast, the masked autoregressive paradigm relaxes the sequential dependency by predicting a subset of masked tokens in parallel, conditioned on the visible ones. 
During training, a random binary mask $M=[m^1,m^2,\ldots,m^N]$ is applied, 
and the model predicts all masked positions with a single forward pass:
\begin{equation}
    \mathcal{L}_{\text{MaskedAR}}
    = -\!\!\!\sum_{i:\,m^i=1}\!\!\!
    \log p_\theta(x_d^i\mid X_d^{\setminus i}),
\end{equation}
where $X_d^{\setminus i}$ denotes the sequence with its $i$-th token masked.

At inference time, the model iteratively refines the masked positions over 
autoregressive steps $t$:
\begin{equation}
    \hat{x}_d^{\,i,(t)} \sim
    p_\theta(x_d^{\,i}\mid \hat{X}_d^{\setminus i, (t-1)}),
\end{equation}
until all tokens are updated. $\hat{x}_d^{\,i,(t)}$ denotes that the generated new tokens at the masked position with $i:\,m^i=1$ at step $t$. $\hat{X}_d^{\setminus i, (t-1)}$ represent the generated sequence from step $t-1$ with $i$-th token masked.

Compared with next-token prediction, masked autoregression naturally aligns with the two-dimensional structure of images: each prediction step conditions on a set of spatially distributed visible tokens instead of a rigid 1D prefix. This flexible conditioning enables the model to better capture local and global spatial dependencies, making it particularly suited for our discrete–continuous mixture formulation. Our MixAR is implemented on top of this paradigm.

% The continuous autoregressive paradigm is first explored by MAR~\cite{mar}, which also belongs to the masked autoregressive paradigm. 
\paragraph{Continuous masked autoregression.} Based on this masked autoregressive paradigm, MAR~\cite{mar} proposed to build continuous autoregression. Given the continuous tokens $X_c = [x_c^1, x_c^2, \ldots, x_c^N]$ from a variational autoencoder (VAE), MAR selects $\lceil r\cdot N\rceil$ tokens and replaces them with a learnable mask token, where $r$ denotes the masking ratio sampled from a pre-defined distribution $p(r)$.  The masked sequence is then fed into a transformer-based model, which predicts the ground-truth latent $x_c^i$ at each masked position $i$ based on the context, following the masked autoregressive paradigm.

% an image $I \in \mathbb{R}^{H\times W\times 3}$, MAR first employs a pre-trained variational autoencoder (VAE) to encode $I$ into latent representations $I'\in\mathbb{R}^{h\times w\times d}$, where $h,w,d$ denote the latent height, width, and channel dimensions respectively.  
% The latent feature map $I'$ is then reshaped into a sequence of $N=h\times w$ latents
% \[
% X_c = [x_c^1, x_c^2, \ldots, x_c^N],
% \]
% where each $x_c^i \in \mathbb{R}^d$ is a continuous vector.  
% During training, $\lceil r\cdot N\rceil$ latents are randomly selected and replaced with a learnable mask token $t_m$, where $r$ denotes the masking ratio sampled from a pre-defined distribution $p(r)$.  
% The masked sequence is then fed into a transformer-based model, which predicts the ground-truth latent $x_c^i$ at each masked position $i$ based on the context, following the masked autoregressive paradigm. 

\paragraph{Diffusion losses.} To effectively handle the high complexity of continuous distributions, MAR introduces a lightweight diffusion head that refines the autoregressive prediction into final latent estimates.  
Formally, given the backbone output $z^i$ at each masked position $i$, the diffusion head learns to denoise a noise-corrupted version of the target latent $x_c^{i,t}$ under the standard denoising objective:
\begin{equation}
\mathcal{L}(z^i, x_c^i) = 
\mathbb{E}_{\varepsilon, t}
\!\left[
  \big\| \varepsilon - \varepsilon_\theta (x_c^{i,t} \mid t, z^i) \big\|^2
\right],
\end{equation}
where $\varepsilon\!\sim\!\mathcal{N}(0,I)$, $t$ denotes the diffusion timestep, and $\varepsilon_\theta$ denotes the diffusion head.  
Although the diffusion loss benefits optimization, continuous autoregressive modeling remains challenging due to the complexity of the representation space, motivating us to further facilitate it with a mixture training mechanism.

\subsection{Factorization Perspective}

Different from MAR, which directly models the joint distribution of continuous latents $p_\theta(X_c)$ in an unconditional manner, our MixAR framework adopts a factorized formulation that leverages an auxiliary discrete representation as an informative prior.  
Specifically, we first employ a pretrained discrete autoregressive model to approximate the discrete joint distribution $p_{\phi}(X_d)$, and then learn a conditional model over continuous latents:
\begin{equation}
p_\theta(X_c \mid X_d) = \prod_{i=1}^{N} p_\theta(x_c^i \mid x_c^{<i}, X_d),
\end{equation}
where $X_d$ provides structured guidance derived from the learned discrete manifold.  
Such guidance offers an explicit prior for the continuous autoregressive model, substantially reducing the complexity of directly modeling the full continuous latent space and making continuous generation more tractable.
% ---------- Figure: DC-Mix vs MAR ----------
\begin{figure}[t]
    \centering
    \includegraphics[width=\columnwidth,trim=18 14 18 6,clip]{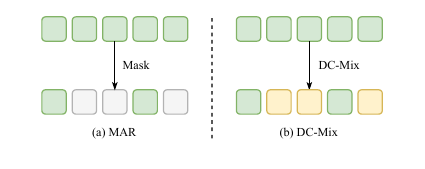}
    \caption{
        \textbf{Comparison between MAR and DC-Mix.}  
        (a) In MAR, masked positions are replaced by a single meaningless mask token before prediction.
        (b) In contrast, DC-Mix pads masked regions with informative discrete tokens that carry semantic and structural cues, making the prediction easier with no additional computation cost. The legend follows the same convention as Figure~\ref{fig:pipeline}.
    }
    \vspace{-3mm}
    \label{fig:dcmix_vs_mar}
\end{figure}

\subsection{Discrete-Continuous Mixture}

The MixAR framework begins with \textbf{Discrete–Continuous Mixture (DC-Mix)}, a novel guidance strategy for injecting discrete conditional signals into continuous modeling. We employ both pretrained variational autoencoder (VAE)~\cite{vavae} and vector-quantized VAE (VQ-VAE)~\cite{maskbit} to extract two complementary representations:
\begin{equation}
\mathbf{x}_c = \text{VAE}(I),\quad
\mathbf{x}_d = \text{VQVAE}(I).
\end{equation}
Here, $\mathbf{x}_c \in \mathbb{R}^{h\times w\times d_c}$ denotes the continuous representations, while $\mathbf{x}_d \in \mathbb{R}^{h\times w\times d_d}$ represents the discrete ones.  

During training, both are flattened into sequences $X_c = [x_c^1, x_c^2, \ldots, x_c^N]$ and $X_d = [x_d^1, x_d^2, \ldots, x_d^N]$, where $N = h\times w$.  
Following the masked autoregressive algorithm, we randomly select $N' = \lceil r \cdot N \rceil$ positions $[k_1, k_2, \ldots, k_{N'}]$ and generate a mask sequence $M=[m^1, m^2, ..., m^N]$ where
\begin{equation}
m^i=
\begin{cases}
1, & i \in [k_1, k_2, \ldots, k_{N'}],\\
0, & \text{otherwise.}
\end{cases}
\end{equation}
As illustrated in Figure~\ref{fig:pipeline}, instead of replacing the masked elements $x_c^i$ with a homogeneous learnable mask token as MAR, we substitute each with its discrete counterpart $x_d^i$ (as shown in Figure~\ref{fig:dcmix_vs_mar}), producing a mixed sequence $\tilde{X} = [\tilde{x}^1, \tilde{x}^2, \ldots, \tilde{x}^N]$ defined by
\begin{equation}
\tilde{x}^i =
\begin{cases}
x_d^i, & m^i = 1,\\
x_c^i, & m^i = 0.
\end{cases}
\label{eq:dcmix}
\end{equation}
The mixed sequence $\tilde{X}$ is then fed into the transformer-based backbone model, which outputs predictions $z^i$ for each masked position.  
Each $z^i$ is used to reconstruct the corresponding $x_c^i$ via a diffusion head, following the MAR setting described in Section~\ref{sec:mar}.  

\paragraph{Comparison with other guidance strategies.} 
We also tried other strategies to inject the guidance, such as the self-attention (DC-SA) and cross-attention (DC-CA). DC-SA takes all discrete tokens $X_d = [x_d^1, x_d^2, \ldots, x_d^N]$ as prefix contextual tokens. In the masked autoregressive process, all discrete tokens are used as seen tokens in self-attention to recover the masked tokens, with the $2\times$ sequence length. While DC-CA takes all discrete tokens $X_d = [x_d^1, x_d^2, \ldots, x_d^N]$ as prefix contextual memory and calculate as cross attention between these discrete tokens and continuous sequences. 
Unlike prefix-based self-attention, which expands the input length from $N$ to $2N$ and thus increases attention complexity from $O(N^2)$ to $O((2N)^2)\!=\!4O(N^2)$, DC-Mix preserves the original sequence length.
Compared with cross-attention, whose theoretical complexity remains $O(N^2)$, DC-Mix still offers practical efficiency gains by avoiding additional projection parameters and extra attention operations required at every block, thereby reducing both computation and memory overhead.  
Despite its lightweight design, we demonstrate that DC-Mix achieves generation fidelity comparable to these more complex guidance strategies, making it an efficient and scalable solution for continuous autoregressive modeling.

% ---------- Figure 2: TI-Mix ----------
\begin{figure}[t]
    \centering
    \includegraphics[width=\linewidth,trim=13 0 13 4,clip]{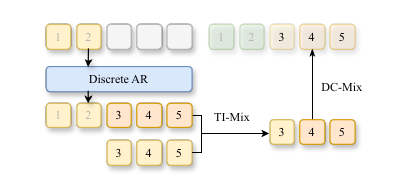}
    \caption{
        \textbf{Illustration of TI-Mix}. As shown in Figure~\ref{fig:dcmix_vs_mar}, without TI-Mix, only ground truth discrete tokens are used as guidance during training, which is different from the generated tokens used in inference. In contrast, TI-Mix uses a mixture of ground truth and generated discrete tokens as guidance during training, effectively mitigating the distribution gap. Here the positions 3,4,5 form the masked region shared by ground truth discrete and continuous tokens. The legend follows the same convention as Figure~\ref{fig:pipeline}.
    }
    \label{fig:timix}
    \vspace{-3mm}
\end{figure}

\subsection{Training-Inference Mixture (TI-Mix)}
When discrete guidance is introduced into continuous modeling, the tokens used at inference time, denoted as $\hat{X}_d=[\hat{x}_d^1,\hat{x}_d^2,\ldots,\hat{x}_d^N]$, are generated by the pretrained discrete AR model rather than taken from ground truth $X_d=[x_d^1,\ldots,x_d^N]$.  
This distributional discrepancy ($p(\hat{X}_d)\!\neq\!p(X_d)$) leads to a training–inference gap. We employ TI-Mix to mitigate this gap.

Given a sequence of discrete tokens $X_d=[x_d^1,x_d^2,\ldots,x_d^N]$, the binary mask $M=[m^1,m^2,\ldots,m^N]$ used for DC-Mix, a pretrained AR model $g_M$ in the masked autoregressive paradigm and its learned mask token $t_m$, we first construct the masked sequence
\begin{equation}
x_d^{\prime i}=
\begin{cases}
t_m, & m^i=1,\\
x_d^i, & m^i=0.
\end{cases}
\end{equation}
Feeding $X_d'=[x_d^{\prime1},\ldots,x_d^{\prime N}]$ into $g_M$ yields the model prediction
\begin{equation}
\hat{X}_d=[\hat{x}_d^1,\hat{x}_d^2,\ldots,\hat{x}_d^N]=g_M(X_d'),
\end{equation}
where $\hat{x}_d^i$ is generated only for masked region(where $m^i=1$) and keeps the same as original $x_d^i$ for the unmasked region(where $m^i=0$).  
Now for the masked region, we have both ground truth and generated discrete tokens. Then TI-Mix is done by partially replace ground-truth tokens in the masked region with generated ones, as shown in ~\ref{fig:timix}  
Let $\lambda\!\in\![0,1]$ denotes the ratio of ground-truth guidance, and a random variable $\rho^i\!\sim\!\mathcal{U}(0,1)$ for each position $i$.  
The final mixed sequence $\tilde{X}_d=[\tilde{x}_d^1,\tilde{x}_d^2,\ldots,\tilde{x}_d^N]$ is defined by
\begin{equation}
\tilde{x}_d^i =
\begin{cases}
\tilde{x}_d^i, & m^i=1~\text{and}~\rho^i<\lambda,\\
\hat{x}_d^i, & m^i=1~\text{and}~\rho^i\ge\lambda,\\
x_d^i, & m^i=0.
\end{cases}
\end{equation}
The tokens in the masked region of $\tilde{X}_d$ serve as new guidance for continuous modeling.  
By monotonically decreasing $\lambda$ during training, the model gradually transitions from ground-truth to generated guidance, improving its adaptation to realistic inference conditions.

% \begin{table}[t]
% \centering
% \small
% \setlength{\tabcolsep}{3.5pt}
% \caption{Quantitative comparisons of Discrete and Continiuous AR models on ImageNet-256. \jiayou{I think we need some descriptive text about this table in Sec4}}
% \vspace{2pt}
% \begin{tabular}{lccccc}
% \toprule
% \textbf{Method} & \textbf{rFID$\downarrow$} & \textbf{Params} & \textbf{gFID$\downarrow$} & \textbf{IS$\uparrow$} & \textbf{Pre./Rec.$\uparrow$} \\
% \midrule
% \multicolumn{6}{c}{\textbf{Discrete AR Models}} \\
% VAR-d30-re~\cite{var} & 1.78 & 2.0B & 1.73 & 350.2 & 0.82 / 0.60 \\
% MaskBit~\cite{maskbit}    & 1.61 & 305M & 1.56 & 312.3 & -- \\
% TiTok      & 1.71 & 287M & 1.97 & 281.8 & -- \\
% RandAR-XXL & 2.19 & 1.4B & 2.15 & 321.97 & 0.79 / 0.62 \\
% MAGVIT-v2  & --   & 307M & 1.78 & 319.4 & -- \\
% LlamaGen-3B~\cite{llamagen} & 0.94 & 3.1B & 2.18 & 263.3 & -- \\
% \midrule
% \multicolumn{6}{c}{\textbf{Continuous AR Models}} \\
% HART-d24   & 0.41 & 1.0B & 2.00 & 331.5 & -- \\
% HART-d30   & 0.41 & 2.0B & 1.77 & 330.3 & -- \\
% ACDIT-H    & 1.22 & 954M & 2.37 & 273.3 & 0.82 / 0.57 \\
% MAR-B~\cite{mar}      & 1.22 & 208M & 2.31 & 281.7 & 0.82 / 0.57 \\
% MAR-L~\cite{mar}      & 1.22 & 479M & 1.78 & 296.0 & 0.81 / 0.60 \\
% MAR-H~\cite{mar}      & 1.22 & 943M & 1.55 & 303.7 & 0.81 / 0.62 \\
% \midrule
% \multicolumn{6}{c}{\textbf{Mixture AR Models}} \\
% MixAR-B    & 0.28 & 208M & 1.99 & 276.96 & 0.61 / 0.67 \\
% MixAR-L    & 0.28 & 557M & 1.53 & 305.99 &  \\
% \bottomrule
% \end{tabular}
% \label{tab:ar_comparison}
% \end{table}

\begin{table}[t]
\centering
\small
\setlength{\tabcolsep}{4pt}
\caption{Quantitative comparisons of Discrete and Continuous AR models on ImageNet-256. Note that the MaskBit results are reproduced under our evaluation protocol.}
\label{tab:main_results}
\vspace{2pt}
\begin{tabular}{lcccc}
\toprule
\textbf{Method} & \textbf{rFID$\downarrow$} & \textbf{\# Params} & \textbf{gFID$\downarrow$} & \textbf{IS$\uparrow$} \\
\midrule
\multicolumn{5}{c}{\textbf{Discrete AR Models}} \\
VAR-d30-re~\cite{var} & 1.78 & 2.0B & 1.73 & 350.2 \\
MaskBit~\cite{maskbit}    & 1.61 & 305M & 1.56 & 312.3 \\
TiTok      & 1.71 & 287M & 1.97 & 281.8 \\
RandAR-XXL & 2.19 & 1.4B & 2.15 & 321.97 \\
MAGVIT-v2  & --   & 307M & 1.78 & 319.4 \\
LlamaGen-3B~\cite{llamagen} & 0.94 & 3.1B & 2.18 & 263.3 \\
\midrule
\multicolumn{5}{c}{\textbf{Continuous AR Models}} \\
HART-d24   & 0.41 & 1.0B & 2.00 & 331.5 \\
HART-d30   & 0.41 & 2.0B & 1.77 & 330.3 \\
ACDIT-H    & 1.22 & 954M & 2.37 & 273.3 \\
MAR-B~\cite{mar}      & 1.22 & 208M & 2.31 & 281.7 \\
MAR-L~\cite{mar}      & 1.22 & 479M & 1.78 & 296.0 \\
MAR-H~\cite{mar}      & 1.22 & 943M & 1.55 & 303.7 \\
\midrule
\multicolumn{5}{c}{\textbf{Mixture AR Models}} \\
MixAR-B    & 0.28 & 208M & 1.99 & 276.96 \\
MixAR-L    & 0.28 & 557M & 1.53 & 305.99 \\
\bottomrule
\end{tabular}
\end{table}
\vspace{-3mm}

%DC-SA+LE
%DC-Mix+VQ
%DC-CA+LE
%在experiments的cite说
%caption里都不说setting
%名字字体align

% efficiency
% DC-Mix放上面

% 不放太多
% Maskbit 1.56
% RAR去掉
% DisCon不写

% CVPR arxihve policy

% Figure4 patch数减小 4*4 T
% 改caption mixture T
% intuition idea

\renewcommand{\subsubsectionautorefname}{Section}

\section{Experiments}
\label{sec:experiments}

\begin{figure*}[t]
    \centering

    % ---------- Left Panel ----------
    \begin{subfigure}[t]{0.49\textwidth}
        \centering
        \includegraphics[width=\linewidth]{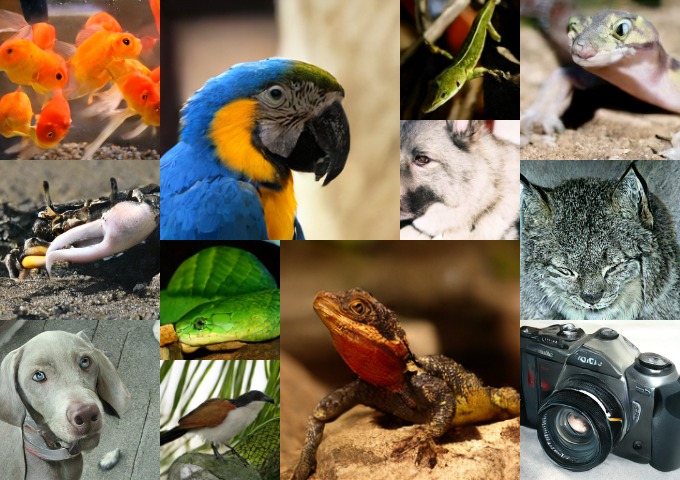} % 左边整张大图
        \caption{MixAR-base} % 可修改
    \end{subfigure}
    \hfill
    % ---------- Right Panel ----------
    \begin{subfigure}[t]{0.49\textwidth}
        \centering
        \includegraphics[width=\linewidth]{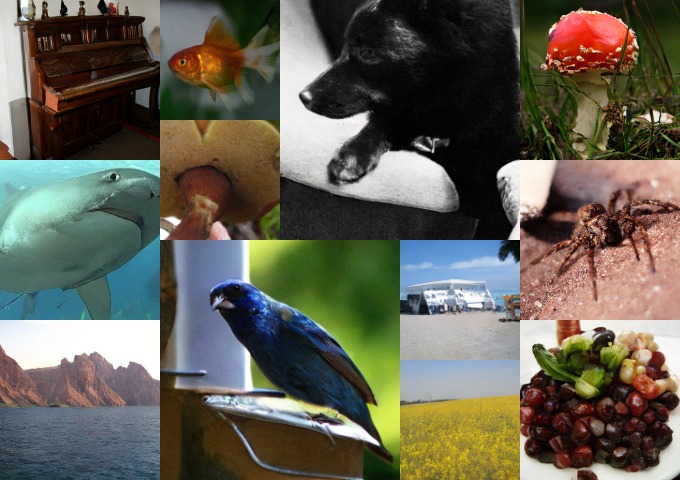} % 右边整张大图
        \caption{MixAR-large} % 可修改
    \end{subfigure}

    \vspace{2mm}
    \caption{
        \textbf{Qualitative Comparison.} 
        Samples generated by MixAR-base (left) and MixAR-large (right). 
        Our results exhibit high-fidelity textures, consistent global structure and diverse semantics across categories.
    }
    \label{fig:qualitative_comparison}
    \vspace{-3mm}
\end{figure*}

We conduct all experiments on the ILSVRC-2012 ImageNet dataset, using the commonly adopted 256×256 resized version (ImageNet-256)~\cite{imagenet}. Each image is pre-tokenized using two complementary tokenizers: a discrete tokenizer from Maskbit~\cite{maskbit} and a continuous tokenizer based on the VA-VAE framework proposed in LightningDiT~\cite{vavae}. We use Maskbit~\cite{maskbit} as our discrete AR model. The architecture of the continuous AR model is adopted from MAR~\cite{mar}.

\subsection{Comparison with MAR}
 A comprehensive quantitative summary is presented in Table \ref{tab:main_results}.
 Across all settings, MixAR demonstrates consistently strong performance compared to both discrete and continuous AR baselines. Notably, MixAR-L surpasses MAR-H in terms of rFID and achieves a slightly better gFID and IS, despite using nearly half the parameters. This highlights the effectiveness of introducing discrete priors to guide continuous autoregressive modeling, improving fidelity without additional computational burden. When compared to discrete AR models such as LlamaGen-3B and MaskBit, MixAR offers substantially lower rFID while maintaining competitive IS scores, demonstrating its ability to capture fine-grained details lost in quantized representations. Moreover, MixAR-B outperforms the similarly sized MAR-B baseline across all metrics, confirming that even lightweight configurations benefit from the proposed mixture formulation. Overall, MixAR achieves a favorable balance between model complexity, fidelity, and efficiency, outperforming larger discrete and continuous counterparts while maintaining parameter efficiency. 

% Across all settings, MixAR achieves consistently strong performance: notably, MixAR-L surpasses MAR-H despite using fewer parameters, demonstrating the effectiveness of incorporating discrete guidance into continuous autoregressive modeling. The overall comparison is summarized in Table~\ref{tab:main_results}.

\subsection{Properties of DC-Mix}
\begin{table}[t]
    \centering
    \small
    \setlength{\tabcolsep}{3pt}
    \caption{
        Efficiency comparison of different guidance mechanisms in the Large model setting with the same backbone and diffusion head.
        % Inference time reports only the continuous model latency, as all settings share the same discrete AR model.
        % All results are obtained on ImageNet-256 with a batch size of 48.
    }
    \vspace{1pt}
    \begin{tabular}{lccc}
    \toprule
    Method & Inference (sec/img) & Train (min/epoch) & Params (M) \\
    \midrule
        DC-Mix      & \textbf{0.031} & \textbf{39.40} & \textbf{557} \\
        DC-SA   & 0.038 & 47.28 & 574 \\
        DC-CA   & 0.037 & 44.55 & 707 \\
    \bottomrule
    \end{tabular}
    \label{tab:efficiency_comparison}
\end{table}

We implement DC-Mix by replacing the learnable mask-token embedding in MAR with the embeddings of padded discrete tokens extracted from the VQ-VAE codebook. For DC-SA and DC-CA, we follow the standard practice of discrete sequence modeling and use learnable discrete token embeddings. DC-SA injects discrete guidance by prepending discrete tokens as prefix inputs to the continuous sequence, while DC-CA introduces guidance by inserting an additional cross-attention block after every self-attention block in the backbone.
Both DC-SA and DC-CA involve 512 tokens, including 256 continuous tokens and 256 discrete ones. In contrast, DC-Mix use 256 mixed tokens with 64 class tokens, reducing the token count by 37.5\%.
All three variants are built on the same base-size continuous autoregressive backbone, with identical architectural depth (same number of attention blocks) and the same diffusion head.
Discrete tokens for both training and inference are obtained using the Maskbit tokenizer and its corresponding generator.

\paragraph{Comparison on efficiency.}
Table~\ref{tab:efficiency_comparison} compares the efficiency of different discrete–continuous guidance mechanisms under the same backbone and diffusion head configuration.
Among the three strategies, DC-Mix demonstrates the best overall efficiency, achieving the fastest inference speed (0.031 s/img) and shortest training time (39.40 min/epoch), while using the fewest parameters (557 M). In contrast, DC-SA and DC-CA require additional attention layers to integrate discrete guidance, which increases both computational cost and model size. Specifically, DC-SA adds a modest overhead, while DC-CA incurs the highest cost. Despite these differences, all methods share the same backbone, confirming that the observed efficiency gain stems solely from the simplicity of DC-Mix’s design.

\begin{figure}[t]
    \centering
    \includegraphics[width=\linewidth,trim=10 5 5 0,clip]{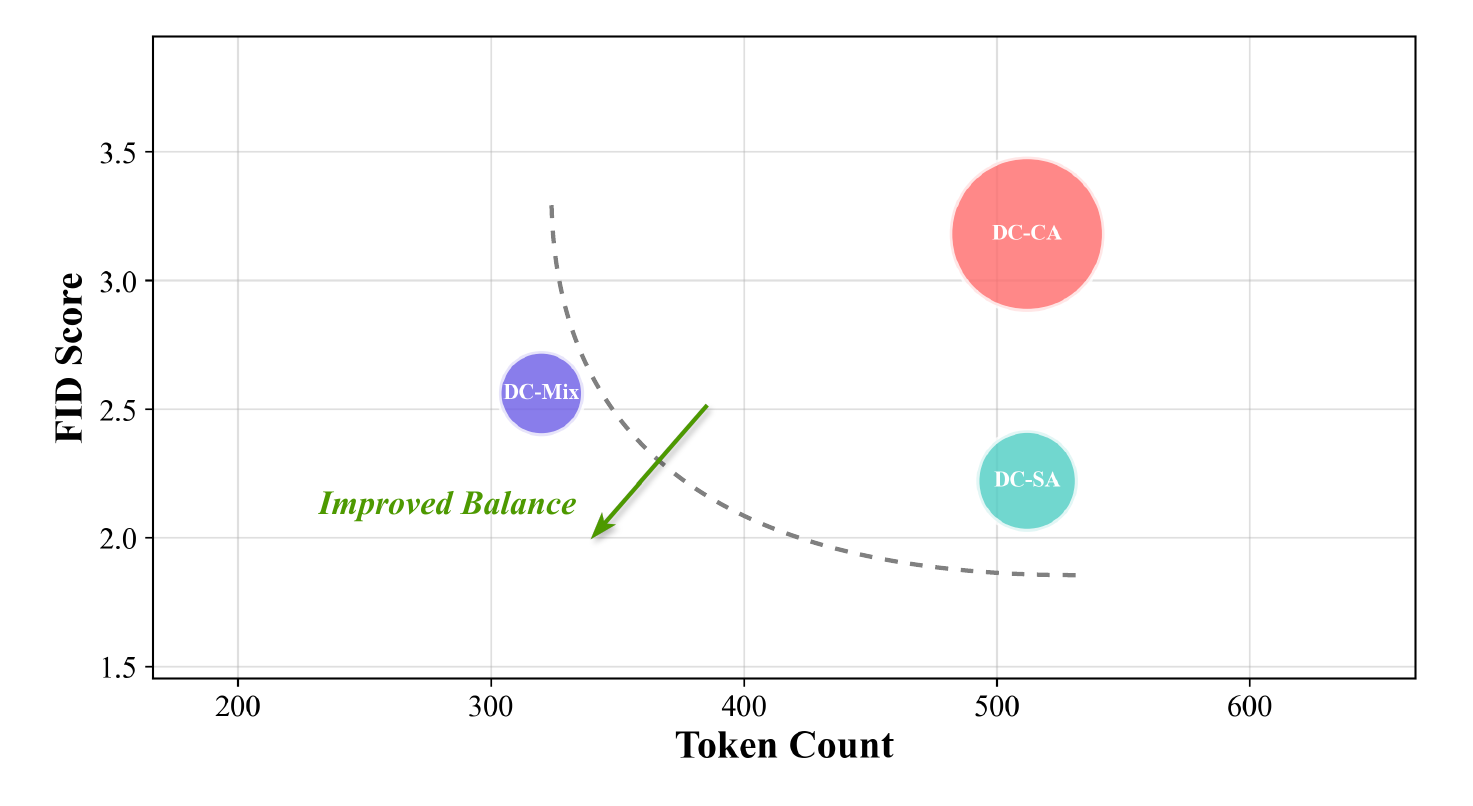}
    \caption{
        \textbf{Balance plot of different guidance strategies.}
        Our DC-Mix achieve comparable performance with DC-SA while reduce token count by a large portion, achieve an ideal balance between efficiency and fidelity. Both DC-SA and DC-Mix achieve higher generation fidelity (low FID score.)
    }
    \label{fig:balance}
    \vspace{-3mm}
\end{figure}

\paragraph{Comparison on the balance of accuracy and efficiency.}

Figure~\ref{fig:balance} presents a balance plot comparing different discrete–continuous guidance strategies in terms of FID score versus token count and model size. The token count on the horizontal axis denotes the total number of tokens used for model prediction, including discrete tokens, continuous tokens, and the CLS token. All results are reported after 320 training epochs on ImageNet-256 under identical optimization settings.
Among the three variants, DC-Mix achieves a markedly better balance between performance and efficiency. It attains a comparable FID score to DC-SA, while using significantly fewer tokens, and far fewer than DC-CA. This indicates that DC-Mix can retain high generation fidelity while reducing the number of tokens required for autoregressive modeling, thereby improving computational efficiency.
The trend shown in the figure highlights that DC-Mix lies closest to the optimal trade-off curve, achieving an ideal balance between model compactness and image quality, whereas DC-SA and DC-CA exhibit diminishing returns with increasing token counts.  In practice, if absolute performance is the priority and a higher training cost is acceptable, DC-SA is the most effective choice, as it leverages all tokens for guidance. For computationally constrained settings, we recommend DC-Mix, which offers a favorable trade-off between efficiency and fidelity.

% In Figure~\ref{fig:balance}, the token count on the horizontal axis denotes the total number of tokens used for model prediction, including discrete tokens, continuous tokens, and the CLS token. All results are reported after 320 training epochs on ImageNet-256 under identical optimization settings.

\subsection{Benefits of TI-Mix}

\begin{figure}[t]
    \centering
    \includegraphics[width=\linewidth,trim=5 5 5 5,clip]{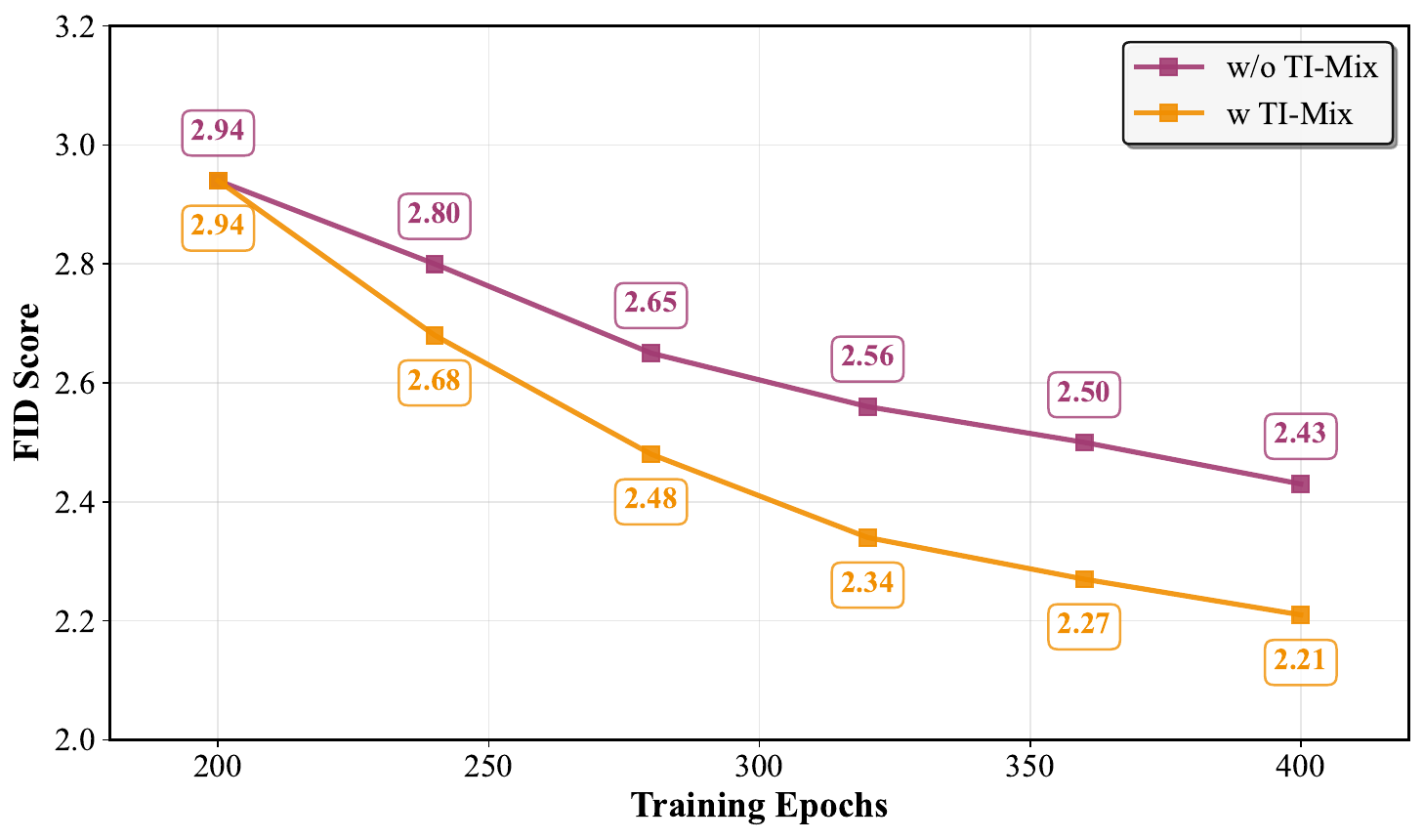}
    \caption{
        \textbf{Ablation Study of TI-Mix.}
        For model trained with 200 epochs DC-Mix, we continue training with another 200 epochs. Extensive training with TI-Mix significantly outperform with simply increase the training epoch nums.
    }
    \label{fig:ti-mix ablation}
    \vspace{-3mm}
\end{figure}

For TI-Mix, we employ a pretrained MaskBit model to generate discrete guidance signals, without introducing any architectural modifications or additional fine-tuning steps. This design choice ensures that TI-Mix can be seamlessly integrated into existing autoregressive pipelines as a plug-and-play enhancement. As illustrated in Figure~\ref{fig:ti-mix ablation}, TI-Mix consistently improves generation fidelity across various training stages. Compared to simply extending training epochs—which often leads to diminishing returns—applying TI-Mix enables the model to better align the training and inference distributions, thereby achieving higher fidelity and more stable convergence. These results demonstrate that incorporating discrete priors through TI-Mix provides a more effective and computationally efficient means of improving continuous autoregressive modeling than prolonged or naive retraining.

% For TI-Mix, we employ pretrained Maskbit for guidance generation, without any architectural modification or fine-tuning. Our experiments demonstrate that TI-Mix serves as a plug-and-play strategy to enhance generation fidelity. Compared with simply increasing training epochs, tuning the model with TI-Mix yields superior results (Fig.~\ref{fig:ti-mix ablation}).

%x: token number
%point: model size
%y: fid
\renewcommand{\subsubsectionautorefname}{Section}

\section{Conclusion}

In this work, we presented MixAR, a unified framework that bridges discrete and continuous autoregressive (AR) modeling through mixture-based training paradigms. By leveraging discrete tokens as prior guidance, MixAR effectively combines the stability of discrete modeling with the expressiveness of continuous latent representations. We explored multiple strategies for integrating discrete and continuous components, including DC-SA, DC-CA, and DC-Mix, and proposed TI-Mix to align the training and inference distributions. Extensive experiments demonstrate that MixAR achieves a strong balance between efficiency and fidelity, and show the improvement to continuous AR models. 
% \textbf{boarder Impact.} Our study advances the understanding of hybrid autoregressive modeling and can potentially improve the fidelity of image generation systems. The proposed techniques could be extended to broader modalities such as video or audio generation, offering more expressive and efficient generative models. However, as with most generative models, MixAR may inadvertently reproduce biases or artifacts present in training data. We encourage careful dataset curation and ethical consideration in applications involving human subjects or sensitive content.

\paragraph{Limitations.} While MixAR successfully integrates discrete priors into continuous autoregressive frameworks, several limitations remain. The current approach still relies on pretrained discrete tokenizers, which may constrain the representation quality. 
Future work may explore joint optimization of the tokenizer and AR model, adaptive mixture strategies, and applications to larger-scale multimodal generation tasks.

\label{sec:conclusion}

{
    \small
    \bibliographystyle{ieeenat_fullname}
    \bibliography{main}
}

% WARNING: do not forget to delete the supplementary pages from your submission 
% \input{sec/X_suppl}

\end{document}